\documentclass[11pt]{article}
\usepackage[final]{acl}
\usepackage{times}
\usepackage{latexsym}
\usepackage[T1]{fontenc}
\usepackage[utf8]{inputenc}
\usepackage{microtype}
\usepackage{inconsolata}
\usepackage{graphicx}
\usepackage{amsmath}
\usepackage{pgfplots}
\usepackage{tikz}
\usetikzlibrary{arrows.meta}
\pgfplotsset{compat=1.18}
\usepackage{hyperref}
\usepackage{caption}
\usepackage{subcaption}
\usepackage{amssymb}
\title{Zonkey: A Hierarchical Diffusion Language Model with Differentiable Tokenization and Probabilistic Attention}
\author{Alon Rozental \ 
\texttt{alonzorz1@gmail.com}}
\date{December 2025}
\begin{document}
\maketitle
\begin{abstract}
Large language models (LLMs) have revolutionized natural language processing, yet they remain constrained by fixed, non-differentiable tokenizers like Byte Pair Encoding (BPE), which hinder end-to-end optimization and adaptability to noisy or domain-specific data. We introduce Zonkey, a hierarchical diffusion model that addresses these limitations through a fully trainable pipeline from raw characters to document-level representations. At its core is a differentiable tokenizer (Segment Splitter) that learns probabilistic beginning-of-sequence (BOS) decisions, enabling adaptive splits that emerge as linguistically meaningful (e.g., word boundaries at spaces, sentence starts at periods) without explicit supervision. This differentiability is enabled by our novel Probabilistic Attention mechanism, which incorporates position-specific existence probabilities to simulate soft masking over theoretically infinite sequences while preserving gradients. Sequences decay probabilistically rather than relying on end-of-sequence tokens, supporting variable-length outputs. Hierarchical levels compress sequences into higher abstractions (e.g., character n-grams to word-like vectors, then sentence-like), with reconstruction via our Denoising Diffusion Mixed Model (DDMM) for stable and efficient denoising in latent space. A Stitcher ensures overlap invariance across segments. Trained end-to-end on Wikipedia, Zonkey generates coherent, variable-length text from noise, demonstrating emergent hierarchies and promising qualitative alignment to data distributions compared to entropy-based learnable tokenizers. Our approach advances toward fully gradient-based LLMs, with potential for better domain adaptation and scalable generation. We release the source code for training and reproducing our experiments.\footnote{The source code for training and reproducing our experiments is publicly available at \url{https://github.com/ARozental/Zonkey}.}
\end{abstract}
\section{Introduction}
The rapid advancement of large language models (LLMs) has transformed fields from machine translation to code generation, driven by Transformer architectures \citep{vaswani2017attention} that excel at capturing long-range dependencies in sequential data. However, foundational components such as tokenization remain a bottleneck: traditional methods like Byte Pair Encoding (BPE) \citep{sennrich2016neural} rely on predefined, rule-based merges that are non-differentiable, leading to out-of-vocabulary (OOV) issues, suboptimal handling of noisy text, and an inability to adapt during end-to-end training. This rigidity is exacerbated in hierarchical models, where lower-level representations (e.g., characters) must be aggregated into higher abstractions (e.g., words or sentences), often with fixed-length assumptions that limit flexibility. Diffusion-based generation, while powerful for images, struggles with text due to discrete tokens, semantic distortions from noise, and fixed-output lengths, as seen in early works like D3PM \citep{austin2021structured}.

Zonkey addresses these challenges with a fully differentiable, hierarchical diffusion framework. Key innovations include Probabilistic Attention for soft variable-length handling, a learnable Segment Splitter acting as an adaptive tokenizer, multi-vector compression with contrastive objectives, DDMM diffusion in latent space, and a differentiable Stitcher for overlap-consistent reassembly. Stitched level-$l$ outputs feed directly into level-$(l+1)$ splitting, enabling recursive hierarchies without length limits.

Empirically, despite being trained on a single GPU with Wikipedia data, Zonkey generates coherent text at the sentence level from noise, with emergent word-level and sentence-level hierarchies and adaptive tokenization. These promising initial results position Zonkey as a step toward scalable, fully differentiable hierarchical language models with potential for longer contexts, higher levels of abstraction, and improved domain adaptation.

In the following sections, we detail Probabilistic Attention (Section~\ref{sec:prob_att}), the Segment Splitter (Section~\ref{sec:splitter}), the Compressor (Section~\ref{sec:compressor}), DDMM Diffusion (Section~\ref{sec:ddmm}), the Stitcher (Section~\ref{sec:stitcher}), the end-to-end training and loss calculations (Section~\ref{sec:training}), and the text generation (Section~\ref{sec:generation}).

\section{Related Work}\label{sec:related}

\subsection{Tokenization and Character-Level Language Models}

Traditional large language models rely on fixed, non-differentiable tokenizers such as Byte Pair Encoding (BPE) \citep{sennrich2016neural}, which can lead to out-of-vocabulary issues and poor handling of noisy or domain-specific text. To mitigate these limitations, character- and byte-level models have been explored, operating directly on raw inputs without subword segmentation. ByT5 \citep{xue2021byt5} extends the T5 architecture to UTF-8 bytes, demonstrating robustness to noise and improved performance on character-level tasks. Similarly, CANINE \citep{clark2021canine} proposes a tokenization-free encoder that processes characters via strided convolutions followed by Transformer layers. More recently, the Byte Latent Transformer (BLT) \citep{pagnoni2024blt} introduces dynamic byte-level patching in a latent space, achieving performance comparable to token-based models at scale while avoiding fixed vocabularies. However, these approaches use static processing of bytes or characters and do not learn adaptive, probabilistic segmentation in an end-to-end differentiable manner. In contrast, Zonkey introduces a fully trainable Segment Splitter that emerges linguistically meaningful boundaries (e.g., words and sentences) from raw characters without explicit supervision.

\subsection{Diffusion Models for Text}

Diffusion models have shown promise for text generation by enabling non-autoregressive sampling and fine-grained control. Early discrete diffusion approaches, such as D3PM \citep{austin2021structured}, apply masked diffusion to categorical tokens. Diffusion-LM \citep{li2022diffusion} shifts to continuous embeddings, improving controllability through gradient-based guidance in latent space. Recent advances include large-scale diffusion LMs like LLaDA \citep{nie2025llada} and TESS 2 \citep{ivi2025tess}, which compete with autoregressive models, as well as energy-based variants \citep{nvidia2025edlm}. Notably, Hierarchical Diffusion Language Models (HDLM) \citep{zhu2025hdlm} introduce semantic scale prediction in a discrete diffusion framework, enabling hierarchical generation. Unlike these works, which typically operate on fixed discrete tokens or lack full hierarchy differentiability, Zonkey combines continuous latent diffusion with a character-level hierarchical pipeline, using probabilistic existence for variable lengths.

\subsection{Soft and Probabilistic Attention Mechanisms}

Standard Transformers employ hard masks for causality and padding, which introduce gradient discontinuities \citep{vaswani2017attention}. To address this, differentiable soft-masked attention \citep{athar2022differentiable} modulates contributions with continuous probabilities, enabling end-to-end optimization over masks. Related probabilistic interpretations of attention have been explored for redundancy reduction \citep{nguyen2022improving}. Zonkey's Probabilistic Attention extends these ideas by incorporating position-specific existence probabilities to handle theoretically infinite sequences softly, preserving gradients while supporting emergent variable-length hierarchies.

Overall, while prior work addresses subsets of these challenges—fixed tokenization bottlenecks, diffusion-based generation, or soft attention—Zonkey uniquely integrates a fully differentiable hierarchical pipeline from raw characters to higher abstractions, with probabilistic mechanisms enabling adaptive tokenization and efficient variable-length diffusion.

\section{The Transformer with Probabilistic Attention}\label{sec:prob_att}
Traditional transformer attention assumes all positions are equally ``real,'' using hard masks for padding, causality, or truncation, which can disrupt gradients and limit flexibility for variable-length modeling~\citep{vaswani2017attention}. In Zonkey, sequences are treated as theoretically infinite, with each position $k$ assigned an existence probability $p_k \in (0,1]$, representing $P(\text{position } k \text{ exists} \mid \text{all prior positions exist})$. These probabilities decay cumulatively, enabling soft truncation during inference when $p_k < \varepsilon$ without explicit EOS tokens.

Probabilistic Attention modulates the raw attention scores $s_{qk} = \frac{\mathbf{Q}_q^\top \mathbf{K}_k}{\sqrt{d}}$ by incorporating existence ratios:

\[
s_{qk}' = s_{qk} + \begin{cases} 
\log\left(\frac{p_k}{p_q}\right) & \text{if } k > q \\
0 & \text{otherwise}
\end{cases}
\]

This adjustment, computed in log-space for numerical stability, effectively multiplies the effect position $k$ has over position $q$ by $P(k \text{ real} \mid q \text{ real}) = p_k / p_q$ for future positions ($k > q$), while assuming $P(k \text{ real} \mid q \text{ real}) = 1$ for past or current positions ($k \leq q$). Low-probability positions thus exert minimal influence on high-probability ones, simulating a soft mask that preserves differentiability. This mechanism is a generalization of traditional hard masking: if cumulative existence probabilities drop sharply from 1 to 0 (or $\epsilon$), it yields equivalent results to conventional masking.

\begin{figure}[htbp]
\centering
\includegraphics[width=0.5\textwidth]{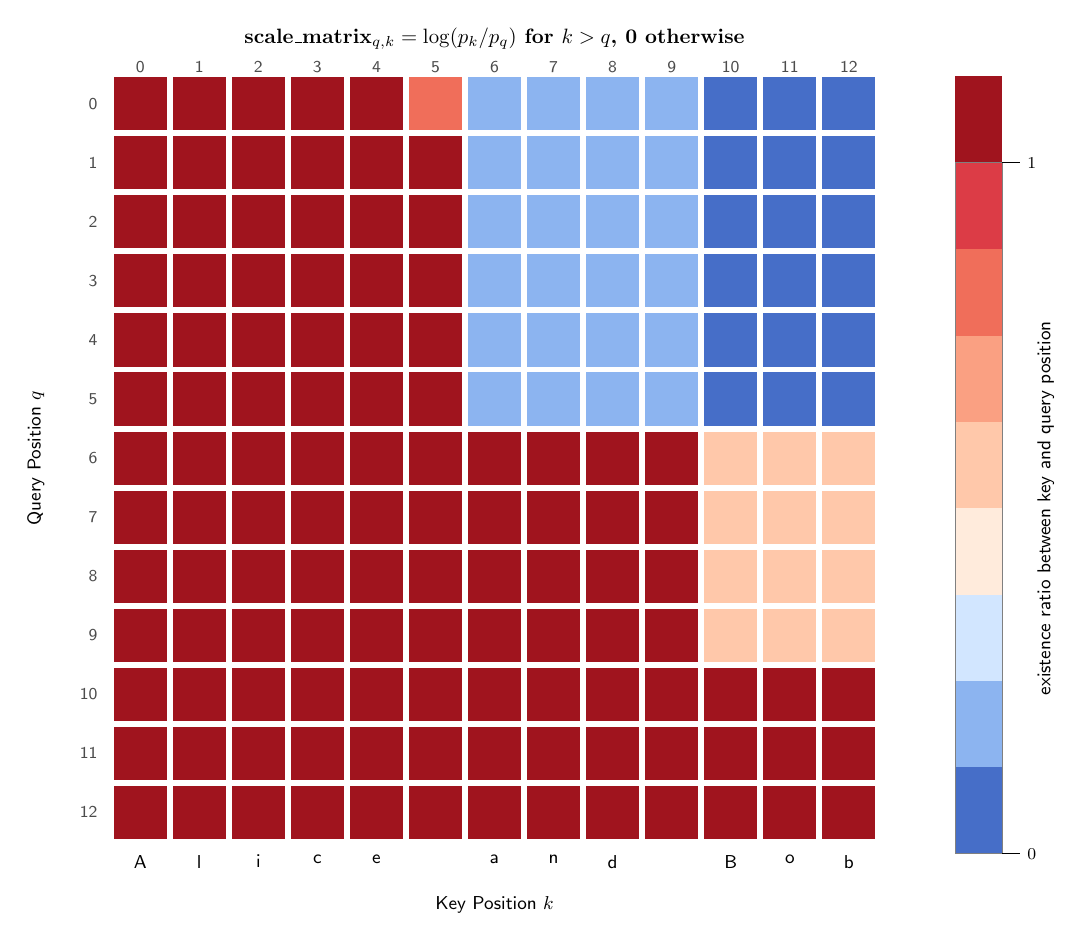}
\caption{Heatmap of the scale matrix adjustment $\log(p_k / p_q)$ in Probabilistic Attention for the example phrase ``Alice and Bob''. Rows correspond to query positions $q$ (top to bottom), columns to key positions $k$. A value of 1 (red) indicates that the attention score is not affected, and this is the case for all $p_q \geq p_k$ positions. Low values (blue) indicate a very small effect of position $k$ over the post-attention state at position $q$. This antisymmetric modulation softly down-weights contributions from low-existence positions, preserving gradients while approximating hard masking.}
\label{fig:prob_att}
\end{figure}

In bidirectional encoders, the mechanism is applied to all position pairs without causality constraints, but scaling occurs only for future positions ($k > q$), downweighting low-probability futures while maintaining neutral influence for past and current positions. This ensures bidirectional contextual integration in hierarchical processing, where existence probabilities propagate from the Splitter and reflect segment uncertainties, focusing modulation on uncertain tails.

In causal decoders (used for autoregressive decompression), the scaling aligns with the masking: future positions receive the adjustment (typically negative for decreasing probabilities), but are already masked to $-\infty$; past and current remain unchanged. During sequential generation, prior positions are assumed to exist with probability 1, aligning with standard decoders; uncertainties in later positions arise from the model's outputs but do not affect allowed attentions due to the masking.

Our model primarily employs Probabilistic Attention in encoders for hierarchical compression, diffusion denoising, and Masked Language Modeling (MLM) tasks, where modulation of future uncertainties stabilizes variable-length reconstructions.

This design is the key enabler for Zonkey's innovations: (1) differentiable splitting in the tokenizer (Splitter), where probabilistic BOS scores propagate gradients through overlaps; (2) hierarchical compression, as higher levels inherit softened attentions from lower ones; and (3) diffusion-based generation in latent space, where noise schedules evolve existence probabilities for variable-length outputs~\citep{austin2021structured,li2022diffusion}. Unlike hard-masked approaches, Probabilistic Attention avoids gradient discontinuities, improving optimization in noisy or uncertain sequences.

Related mechanisms include probabilistic interpretations of attention as Gaussian mixtures~\citep{nguyen2022improving} or maximum a posteriori estimation~\citep{gabbur2021probabilistic}, which address rank collapse or redundancy in heads. Our approach extends these by explicitly conditioning on existence for infinite-sequence modeling.

\section{The Differentiable Tokenizer: Segment Splitter}\label{sec:splitter}

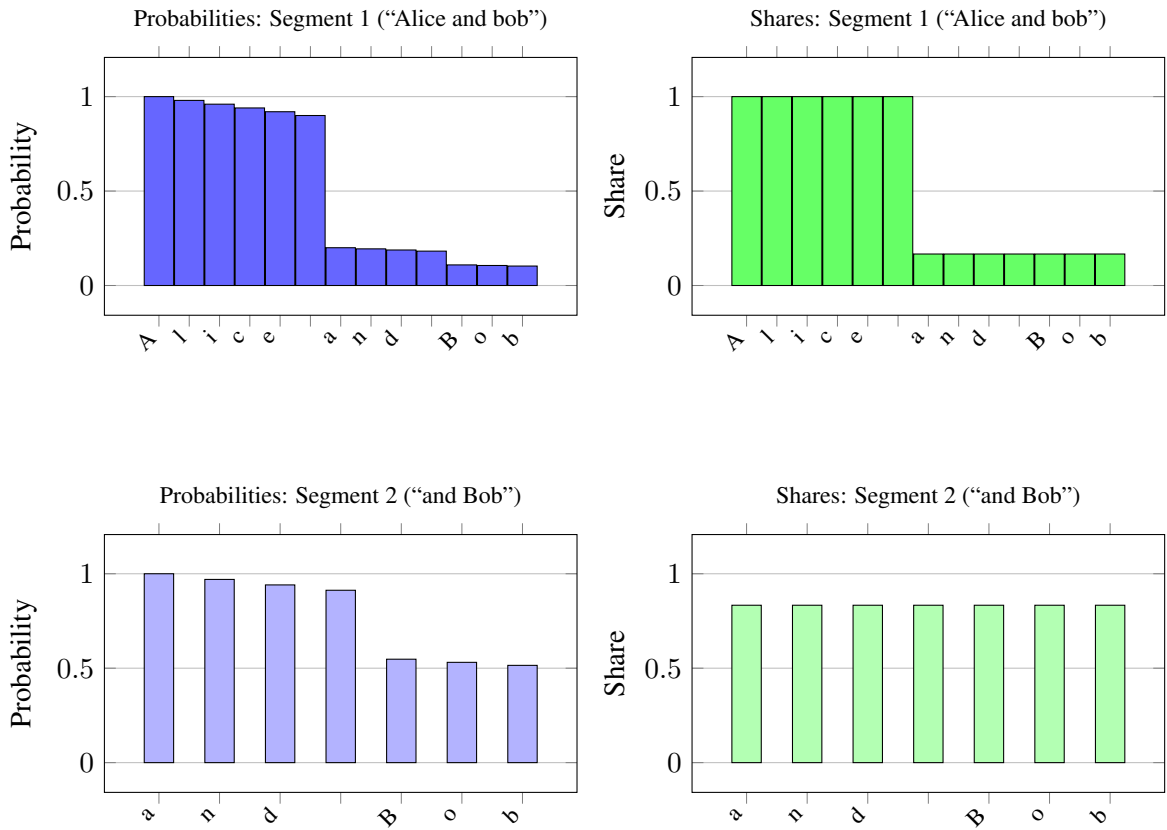
\begin{figure*}
\centering
\begin{tikzpicture}

\def\separation{180} % Vertical spacing between top and bottom
\def\horizontal{220} % Horizontal spacing between left and right columns
\def\horizontalshift{0}
\def\graphwidth{7.8cm} % Slightly wider now that legend is removed
\def\graphheight{5.0cm} % Slightly taller for better visibility

% Top-left: Probabilities Segment 1
\begin{axis}[
    ybar,
    enlargelimits=0.15,
    ylabel={Probability},
    title={Probabilities: Segment 1 (``Alice and bob'')},
    xtick={0,1,2,3,4,5,6,7,8,9,10,11,12},
    xticklabels={A,l,i,c,e, ,a,n,d, ,B,o,b},
    xticklabel style={rotate=45, anchor=east, font=\footnotesize},
    title style={font=\small},
    ymin=0, ymax=1.05,
    ymajorgrids=true,
    bar width=11pt,
    width=\graphwidth,
    height=\graphheight,
    at={(\horizontalshift pt,\separation pt)},
]

\addplot[fill=blue!60] coordinates {
(0,1.0000) (1,0.9800) (2,0.9600) (3,0.9400) (4,0.9200) (5,0.9000) 
(6,0.2000) (7,0.1940) (8,0.1880) (9,0.1820) (10,0.1090) (11,0.1060) (12,0.1030)
};

\end{axis}

% Bottom-left: Probabilities Segment 2
\begin{axis}[
    ybar,
    enlargelimits=0.15,
    ylabel={Probability},
    title={Probabilities: Segment 2 (``and Bob'')},
    xtick={0,1,2,3,4,5,6,7,8,9,10,11,12},
    xticklabels={ , , , , , ,a,n,d, ,B,o,b},
    xticklabel style={rotate=45, anchor=east, font=\footnotesize},
    title style={font=\small},
    ymin=0, ymax=1.05,
    ymajorgrids=true,
    bar width=11pt,
    width=\graphwidth,
    height=\graphheight,
    at={(\horizontalshift pt,0 pt)},
]

\addplot[fill=blue!30] coordinates {
(6,1.0000) (7,0.9700) (8,0.9409) (9,0.9126) (10,0.5476) (11,0.5310) (12,0.5150)
};

\end{axis}

% Top-right: Shares Segment 1
\begin{axis}[
    ybar,
    enlargelimits=0.15,
    ylabel={Share},
    title={Shares: Segment 1 (``Alice and bob'')},
    xtick={0,1,2,3,4,5,6,7,8,9,10,11,12},
    xticklabels={A,l,i,c,e, ,a,n,d, ,B,o,b},
    xticklabel style={rotate=45, anchor=east, font=\footnotesize},
    title style={font=\small},
    ymin=0, ymax=1.05,
    ymajorgrids=true,
    bar width=11pt,
    width=\graphwidth,
    height=\graphheight,
    at={(\horizontal pt + \horizontalshift pt,\separation pt)},
]

\addplot[fill=green!60] coordinates {
(0,1.0000) (1,1.0000) (2,1.0000) (3,1.0000) (4,1.0000) (5,1.0000) 
(6,0.1667) (7,0.1667) (8,0.1667) (9,0.1667) (10,0.1667) (11,0.1667) (12,0.1667)
};

\end{axis}

% Bottom-right: Shares Segment 2 (no legend anymore)
\begin{axis}[
    ybar,
    enlargelimits=0.15,
    ylabel={Share},
    title={Shares: Segment 2 (``and Bob'')},
    xtick={0,1,2,3,4,5,6,7,8,9,10,11,12},
    xticklabels={ , , , , , ,a,n,d, ,B,o,b},
    xticklabel style={rotate=45, anchor=east, font=\footnotesize},
    title style={font=\small},
    ymin=0, ymax=1.05,
    ymajorgrids=true,
    bar width=11pt,
    width=\graphwidth,
    height=\graphheight,
    at={(\horizontal pt + \horizontalshift pt,0 pt)},
]

\addplot[fill=green!30] coordinates {
(6,0.8333) (7,0.8333) (8,0.8333) (9,0.8333) (10,0.8333) (11,0.8333) (12,0.8333)
};

\end{axis}

\end{tikzpicture}
\caption{Illustration of existence probabilities and normalized existence shares
in the Segment Splitter for overlapping segments from the example phrase ”Alice
and Bob”. Left panels: Probabilities for Segment1(full phrase, decaying sharply
after ”Alice ”) and Segment 2 (starting at ”and Bob”). These are derived from
cumulative products of (1 - p BOS), simulating infinite sequences with natural
truncation at low values. Right panels: Corresponding normalized shares for
loss calculation, ensuring each position contributes uniformly (summing to 1
across overlaps) despite redundancy in positions 6-12. This mechanism preserves
differentiability and incentivizes semantically meaningful splits.}
\label{fig:splitter_overlap}
\end{figure*}

The Segment Splitter serves as Zonkey's hierarchical tokenizer, transforming an input sequence—such as character embeddings at level 0—into overlapping segments that form the basis for higher-level abstractions (e.g., words at level 0, sentences at level 1). Unlike fixed subword tokenizers like Byte Pair Encoding (BPE)~\citep{sennrich2016neural}, which rely on predefined rules and lack end-to-end differentiability, the Segment Splitter learns probabilistic beginning-of-sequence (BOS) decisions directly from data. This enables adaptive, context-aware splitting that propagates gradients through the hierarchy, addressing limitations in static tokenization schemes~\citep{gong2019limitations}. By integrating with Probabilistic Attention, the splitter incentivizes semantically meaningful splits without explicit supervision, resulting in emergent behaviors such as elevated BOS probabilities for spaces at level 0 (demarcating words) and after periods at level 1 (initiating sentences).

The Splitter takes a sequence of vectors as input, representing tokens from a lower level; for the first pass at level 0, these are the embedding representations of the original document characters during training. At this stage, the input consists solely of these vectors, without existence probabilities yet incorporated. The process begins by computing a BOS probability, \( p_{\text{BOS}, i} \in (\varepsilon, 1 - \varepsilon) \) (where \( \varepsilon \) is a small constant for stability, e.g., \( 10^{-6} \)), for each position \( i \) in the input sequence. These probabilities are generated using a lightweight linear transformer encoder~\citep{beltagy2020longformer}), which processes the input embeddings. We transform the output to probability using a projection with sigmoid activation. This encoder captures local context, such as character n-grams or punctuation patterns, to predict split points.

During training, BOS positions are sampled stochastically: for each position \( i > 0 \), a BOS is selected with probability \( p_{\text{BOS}, i} \); the first position in a document always starts a segment. This hard sampling introduces non-differentiability, as gradients cannot flow through discrete choices~\citep{jang2017categorical}. Ideally, with infinite compute, we would enumerate all \( 2^{L-1} \) possible split configurations for a sequence of length \( L \), weighting each forward pass by its split probability in the loss computation. In practice, this is intractable, so we leverage the raw BOS probabilities within downstream Probabilistic Attention modules. By modulating attention weights with existence ratios derived from these probabilities, the model incurs higher reconstruction and compression losses for suboptimal splits (e.g., those that make downstream MLM and sequence reconstruction harder). This creates an implicit gradient signal: to minimize overall loss, the splitter must learn ``good'' splits that facilitate effective encoding and decoding, effectively backpropagating through the probabilistic framework. Note that alternatives like Gumbel-Softmax~\citep{jang2017categorical} would not suffice here, as the hard choice of which positions indicate the start of a new segment for the rest of the network is not averted.

During inference, BOS decisions are made deterministically by thresholding \( p_{\text{BOS}, i} > 0.5 \). During training the Splitter outputs:

1. \textbf{Segments}: Overlapping subsequences of vectors starting at sampled BOS positions; theoretically, all these segments continue until the end of the document regardless of their starting position. In practice, for computational efficiency we set the length of each segment to exactly \( \text{max\_seq\_len}[l] \)  (e.g., 32); we note that these segments do overlap and segments that start near the end of the document have padding vectors appended to them.

2. \textbf{Per-position BOS probabilities}.

3. \textbf{Existence probabilities}: For each position \( j \) in a segment starting at \( i \), 
\( p_{\text{exist}, j} = \prod_{k=i+1}^{j} (1 - p_{\text{BOS}, k}) \), 
simulating infinite sequences with decaying probability. These are derived from the per-position 
BOS probabilities and enable soft truncation and gradient flow through low-uncertainty splits.

4. \textbf{Average BOS probability}: The mean BOS probability over all positions in a training batch, only used for loss calculation.

5. \textbf{Long segment probability}: This is the average probability over all positions in a training batch that position $i$ starts a patch of at least \( \text{max\_seq\_len}[l] \) with no chosen BOS in it. It is only used for loss calculation and usually has a negligible effect as  \( \text{max\_seq\_len}[l] \)  is set in such a way that the residual existence probabilities after it are very low. 

6. \textbf{Short segment probability}: This is the average probability over all positions in a training batch that position $i$ starts a segment and has another segment starting shortly after. "short" is defined by a hyper-parameter called \( \text{num\_compression\_vectors}[l] \), which is discussed in the next section. It is only used for loss calculation.

7. \textbf{Existence shares}: For each position in the original sequence, a per-segment weight tensor where the position appears, normalized such that shares sum to 1 across all overlapping segments containing it. Formally, for a position \( j \) in segment \( s \), the raw share is the cumulative product of \( 1 - p_{\text{BOS}, k} \) for \( k \) from the segment start to \( j \) (i.e., the probability the segment has not terminated by \( j \)); these are then divided by the sum of raw shares for \( j \) over all segments including it. These shares reweight per-position losses downstream, ensuring uniform contribution regardless of overlap density. This prevents exploitation where the splitter concentrates BOS in predictable text regions while sparsely covering complex areas, which would otherwise bias learning toward easier subsequences. Existence shares are closely tied to existence probabilities, as both are derived from the same cumulative products of \(1 - p_{\text{BOS}, k}\). When \(p_{\text{BOS}, j}\) is high at position \(j\), it simultaneously lowers the existence probabilities for all subsequent positions in any overlapping segment that includes \(j\) and reduces the expected raw (and thus normalized) existence shares for those positions. Existence shares are only used for loss calculation.

These outputs feed into the main hierarchical level, where segments are compressed, noised, and reconstructed. Empirically, training on datasets like Wikipedia yields splits aligning with linguistic structures: at level 0, spaces (or characters following spaces) exhibit \( p_{\text{BOS}} \) spikes due to their role in word boundaries; at level 1, periods and other sentence delimiters similarly peak, emerging from loss minimization rather than hardcoded rules. This adaptability handles noisy or domain-specific text effectively, and contrasts with prior learnable approaches like Byte Latent Transformer~\citep{pagnoni2024blt}, which enforce equal entropy per patch—a suboptimal strategy, as it fails to align with global objectives by wasting compute on high-entropy substrings that are inherently unguessable such as passwords that found their way to the training set and to a lesser extent on unfamiliar proper nouns without incentivizing meaningful compression.

\section{The Compressor: Hierarchical Sequence Compression}\label{sec:compressor}
Following the probabilistic segmentation of input sequences by the Segment Splitter, Zonkey employs a dedicated compression module to distill each (overlapping) segment of length $\text{max\_seq\_len}[l]$ (a hyper-parameter, e.g., 32) but with varying effective lengths determined by the areas of non-negligible existence probabilities—into a fixed-dimensional vector representation suitable for higher-level abstractions. This process is pivotal for building the model's hierarchical structure, where lower-level sequences (e.g., character embeddings at level 0) are aggregated into compact vectors representing semantically richer units (e.g., words or phrases). Unlike traditional pooling or averaging techniques in hierarchical transformers \cite{pappagari2019hierarchical}, our Compressor leverages Probabilistic Attention to handle uncertain segment lengths differentiably, enabling end-to-end optimization across levels. This section details the compression mechanism, including its integration with existence probabilities.
\subsection{Compression Pipeline}
For a given document at hierarchical level $l$, the Splitter produces $m_l$ overlapping segments, each of shape $(\text{max\_seq\_len}[l], d_{\text{model}}[l])$, where $m_l$ is typically much smaller than the original sequence length due to adaptive splitting. These segments are augmented with per-position BOS probabilities and existence shares for weighted loss computation.
To compress a segment, we prepend $N = \text{num\_compression\_vectors}[l]$ (a hyper-parameter, e.g., 4) learnable classification (CLS) vectors to the input. These CLS vectors, initialized as trainable parameters of dimension $d_{\text{model}}[l]$, serve as summarization anchors, similar to BERT's [CLS] token \cite{devlin2019bert} but extended for multi-vector compression to capture richer latent structures. The existence probabilities for these prepended vectors are fixed at 1, ensuring they fully attend to the segment while the segment's existence probabilities decay based on the cumulative multiplication of $1 - p_{\text{BOS}}$.
The augmented sequence is then processed by a multi-layer Transformer Encoder, termed the Compressor, which applies self-attention with Probabilistic Attention to modulate influences from low-probability "tail" positions. Formally, for input sequence $\mathbf{X}$ (after prepending CLS vectors) and existence probabilities $\mathbf{p}$ (1 for CLS, decaying for the segment), the Compressor computes:
$$\mathbf{H} = \text{TransformerEncoder}(\mathbf{X}, \mathbf{p}),$$
where $\mathbf{H}$ is the final hidden states. The compressed representation $\mathbf{c}$ is extracted as the first $N$ vectors from $\mathbf{H}$, flattened to $(N \cdot d_{\text{model}}[l])$, and normalized to match the expected norm of a normally distributed vector of the same length for stability in diffusion processes, which prevents magnitude explosion during training. This setup enables gradients to propagate through uncertain split decisions, incentivizing the Segment Splitter to produce segments that are highly compressible. Ambiguous BOS probabilities effectively inflate the number of plausible token-like units (increasing the ``vocabulary'' cardinality), which in turn elevates downstream MLM and reconstruction losses. Consequently, the model learns to favor crisp, decisive splits that align with linguistically natural boundaries—such as complete words rather than arbitrary character subsequences—consistent with information-theoretic principles showing that meaningful units like words exhibit lower redundancy and entropy \citep{shannon1951prediction}.

\subsection{MLM Loss for Semantic Alignment}
Inspired by BERT \cite{devlin2019bert}, we mask 15\% of non-padding positions in the segment (selected randomly but biased toward high-existence), replacing with a learnable [MASK] token. The Compressor processes this masked input, and a linear head predicts the original vectors at masked positions via cross-entropy over a contrastive set which is composed of the original vector and negatives example that are sampled from the level $l$ vectors of the batch. This loss clusters similar meanings in latent space \cite{gao2021simcse},  ensuring compressed vectors capture semantics: e.g., substituting "excellent" with "great" should incur lower loss than substituting with "cow," as the former preserves semantic intent. This is more readily achieved via MLM's focus on contextual prediction than through cross-entropy on noise-reconstructed vectors alone, where lexical mismatches might dominate without explicit semantic alignment, ensuring higher-level vectors prioritize meaning over surface form for robust hierarchical generation and denoising.

The MLM is weighted by the existence shares of the targets and is omitted at level 0, as predicting individual characters does not advance semantic clustering, hierarchical incentives, or meaning preservation—character-level guessing rarely requires deep contextual understanding, unlike word- or sentence-level infilling.
Empirically, this compression yields vectors that, when decompressed (as described in later sections), reconstruct coherent text hierarchies. The integration with Probabilistic Attention ensures gradients flow back to the Splitter and the original character embedding matrix, enabling adaptive tokenization that theoretically handles domain shifts or noisy data more robustly than static methods \cite{hofstatter2021efficiently}. This positions Zonkey as a step toward fully learnable, gradient-based hierarchies, addressing gaps in prior variable-length models \cite{pagnoni2024blt} by aligning compression directly with downstream reconstruction quality rather than heuristic entropy targets.

\section{DDMM Diffusion: Denoising Diffusion Mixed Model for Hierarchical Latent Reconstruction}\label{sec:ddmm}
Zonkey employs a diffusion-based mechanism to reconstruct Splitter-derived segments from their hierarchical compressions, enabling robust sequence recovery and variable-length generation. This approach, termed Denoising Diffusion Mixed Model (DDMM), integrates the cautious, variance-exploring stochasticity of Denoising Diffusion Probabilistic Models (DDPM; \cite{ho2020denoising}) with the bold, trajectory-efficient determinism of Denoising Diffusion Implicit Models (DDIM; \cite{song2021denoising}). DDPM's small-step reversals foster diversity but lead to inefficient inference (thousands of iterations) and instability in text, where incremental noise accumulates semantic distortions and fixed-length constraints stifle natural hierarchies \cite{austin2021structured,he2023diffusion}. DDIM's larger leaps accelerate sampling but heighten risks of mode collapse and oversmoothing in unanchored latents, particularly for variable-length language where paths diverge unpredictably \cite{gong2023diffuseq}. DDMM bridges these by training the denoiser to handle both paradigms: it rewards safe, small moves when uncertain (emulating DDPM's resilience to partial progress) while encouraging ambitious leaps toward cleans when feasible (mirroring DDIM's directness). The model therefore dynamically calibrates step size based on latent confidence, yielding stable, high-fidelity hierarchies that outperform baselines in semantic coherence and adaptability.
At hierarchical level $l$, DDMM inverts compressions $\mathbf{c}$ (shape $(m_l, N \cdot d[l])$, $N$ compression vectors) to approximate original segments. The pipeline begins by reshaping $\mathbf{c}$ into $N$ lower-level vectors, simulating lower-level "tokens" for recursive processing. Noise injects post-reshaping via variance-preserving: $\tilde{\mathbf{c}} = \sqrt{t} \cdot \boldsymbol{\epsilon} + \sqrt{1 - t} \cdot \mathbf{c}$, $\boldsymbol{\epsilon} \sim \mathcal{N}(0, \mathbf{I})$, preserving norms for embedding stability \cite{sohl2015deep,karras2022elucidating}. 

A noise-conditioning vector \(\mathbf{v}_t = (1 - t) \cdot \mathbf{v}_{\text{clean}} + t \cdot \mathbf{v}_{\text{noisy}}\) is prepended to the $N$ compression vectors,  these serves as "prompt" vectors and are assigned an existence probability of 1. A lightweight decoder (typically 1-2 layers) expands this "prompt" autoregressively into up to $\text{max\_seq\_len}[l]$ positions, yielding sequence which is sufficient for a rough existence probability estimation but not for final vector representations. We aim to keep the vast majority of our free parameters in bidirectional encoders as they better captures context \cite{peebles2023scalable}. This preliminary output informs a BOS classifier, predicting per-position BOS probabilities to derive existence probabilities $\mathbf{p}$, which decay cumulatively as $p_j = \prod_{k=1}^{j} (1 - p_{\text{BOS},k})$.
The noisy sequence, augmented with existence probabilities, undergoes refinement via a multi-layer Transformer Encoder using Probabilistic Attention. This step outputs the denoised sequence $\hat{\mathbf{X}}$, weighted by existence shares for loss computation. 

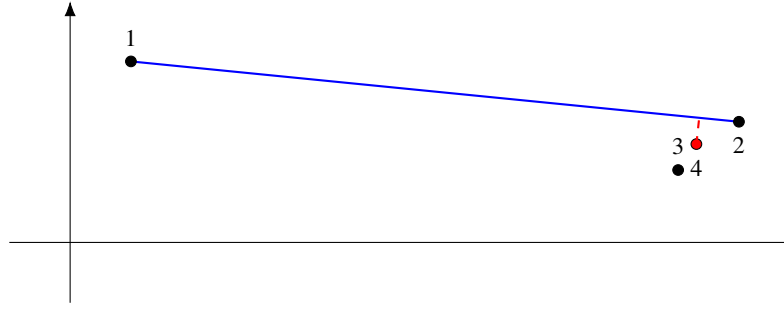
\begin{figure}[h]
\centering
\begin{tikzpicture}[
    scale=0.8,
    point/.style={circle, draw=black, fill=black, inner sep=0pt, minimum size=4pt},
    redpoint/.style={circle, draw=black, fill=red, inner sep=0pt, minimum size=4pt},
    label/.style={font=\small, black}
]
    % Coordinates
    \def\pOneX{1}   \def\pOneY{3}
    \def\pTwoX{11}  \def\pTwoY{2}
    \def\pThreeX{10}\def\pThreeY{1.2}
    \def\pFourX{10.3}\def\pFourY{1.63}
    \def\pFiveX{10.35}\def\pFiveY{2.1}

    % Axes
    \draw[-{Latex[length=2mm]}] (-1,0) -- (12,0);
    \draw[-{Latex[length=2mm]}] (0,-1) -- (0,4);

    % Blue line (training target)
    \draw[blue, thick] (\pOneX,\pOneY) -- (\pTwoX,\pTwoY);

    % Points
    \node[point] (p1) at (\pOneX,\pOneY) {};
    \node[label, above=2pt] at (p1) {1};

    \node[point] (p2) at (\pTwoX,\pTwoY) {};
    \node[label, below=2pt] at (p2) {2};

    \node[point] (p3) at (\pThreeX,\pThreeY) {};
    \node[label, above=2pt] at (p3) {3};

    \node[redpoint] (p4) at (\pFourX,\pFourY) {};
    \node[label, below=2pt] at (p4) {4};

    % Red dashed deviation
    \draw[red, thick, dashed] (\pFourX,\pFourY) -- (\pFiveX,\pFiveY);
\end{tikzpicture}
\caption{The DDMM mixed-step training objective.
\(\mathbf{p}_{1}\): clean compression. 
\(\mathbf{p}_{2}\): heavily noised. 
\(\mathbf{p}_{3}\): denoising and compressing \(\mathbf{p}_{2}\), then add light noise. 
\(\mathbf{p}_{4}\): denoising and compressing \(\mathbf{p}_{3}\). 
Loss is determined by the cosine similarity from \(\mathbf{p}_{4}\) to its projection on the blue segment \([\mathbf{p}_{1},\mathbf{p}_{2}]\)], even though this image uses euclidean distances to ease visualization.}
\label{fig:ddmm_process}
\end{figure}

In addition to the standard final-step denoising loss (predicting the clean latent from a lightly noised version), DDMM introduces a mixed-step objective that encourages larger, more deterministic leaps. Starting from a clean compression \(\mathbf{p}_{1}\), we add substantial noise to obtain \(\mathbf{p}_{2}\). The denoiser and compressor are applied to \(\mathbf{p}_{2}\), producing an intermediate compression; a small amount of noise is then added, simulating one forward step, yielding \(\mathbf{p}_{3}\). The denoiser and compressor are applied once more to \(\mathbf{p}_{3}\), resulting in \(\mathbf{p}_{4}\). The loss minimizes the cosine distance between \(\mathbf{p}_{4}\) and its nearest point on the line segment \([\mathbf{p}_{1}, \mathbf{p}_{2}]\). We use an \(\mathrm{atanh}\) transformation and in-batch negatives, similarly to how we compute our MLM compressor loss. Because there is no penalty for moving in the ``right'' direction, towards \(\mathbf{p}_{1}\), this trains the model to recover from large deviations in a single ambitious step when possible. In other words, when the original text is identifiable from the noisy version, the model is encouraged to move from \(\mathbf{p}_{3}\) towards \(\mathbf{p}_{1}\) rather than \(\mathbf{p}_{2}\) because \(\mathbf{p}_{1}\)'s location is unaffected by randomness. However, in cases where the model is unable to identify \(\mathbf{p}_{1}\) with a large enough measure of certainty, it is encouraged to only take a small step from \(\mathbf{p}_{3}\) and still remain close to \(\mathbf{p}_{2}\) rather than gamble on a large step in the hope of approaching \(\mathbf{p}_{1}\). See Figure~\ref{fig:ddmm_process} for more details.

\section{The Segment Stitcher: Differentiable Reassembly for Hierarchical Consistency}\label{sec:stitcher}
The Segment Stitcher serves as the symmetric counterpart to the Segment Splitter in Zonkey. It reassembles overlapping, denoised (or generated) segments into coherent, variable-length document-level representations while remaining fully differentiable. Traditional long-context models typically rely on hard chunking or naive concatenation, which can introduce boundary artifacts and disrupt gradient flow across segments~\citep{song2024hierarchical,bai2024longwriter}. By contrast, Zonkey's overlapping segmentation paired with a fully differentiable Stitcher ensures seamless information propagation, enforces near-identical representations in overlap regions, and delivers direct supervisory signals that refine upstream components—particularly the probabilistic BOS decisions of the Splitter. This closes the end-to-end optimization loop over arbitrary document lengths, enabling stable hierarchical abstraction and coherent diffusion-based generation of unbounded text.

In practice, stitching is usually straightforward: the Splitter's near-deterministic BOS probabilities produce sharp drops in existence probabilities, making each segment's effective length highly predictable. The primary alignment signal is thus the existence-probability decay in the preceding segment, which reliably indicates where to truncate its tail and append the next. However, occasional denoising imperfections in overlap regions create valuable refinement opportunities. By softly aligning consecutive segments and blending their representations, the Stitcher can correct subtle errors using the complementary view provided by the overlap—updating embeddings even in the non-tail portions of the preceding segment via constrained cross-attention.

Consider a level-1 reconstruction of the sentence ``Quantum computing harnesses quantum mechanics to perform computations exponentially faster than classical computers.'' The first denoised segment might read ``Quantum computing harnesses classical mechanics to perform computations much slower than modern computers.'' with a sharp existence-probability drop after ``to perform''. The second segment, starting later, reconstructs as ``quantum mechanics to solve complex problems efficiently.'' The pronounced existence drop already signals unreliability after ``to perform'' in the first segment. More importantly, the overlap allows the Stitcher to refine the shared region: cross-attention from the second segment updates embeddings in the first, favoring ``quantum mechanics'' and ``exponentially faster'' over the erroneous ``classical mechanics'' and ``much slower''. This gentle correction improves coherence without overwriting the primary denoising work.

\subsection{Stitching Pipeline}
At level $l$, the Stitcher receives $m_l$ denoised segments—each of shape $(\text{max\_seq\_len}[l], d_{\text{model}}[l])$—along with their associated existence probabilities $\mathbf{p}_{\text{exist}}$. The pipeline consists of three fully differentiable stages:

\textbf{Soft Offset Inference.} Pairwise similarities between consecutive segments are computed in a reduced-dimensional projection space ($d_{\text{model}}/4$). These are combined with cumulative existence-probability differences from the preceding segment to yield soft alignment scores. The resulting offset—indicating where the second segment begins within the first—is a continuous, weighted estimate rather than a discrete argmax. In most cases the existence drop dominates, with content similarity providing modest but useful refinement when the overlap contains informative patterns.

\textbf{Weighted Accumulation with Probabilistic Attention Guidance.} Contributions from overlapping positions are blended using existence probabilities. Low-existence tails are naturally down-weighted, ensuring smooth transitions without abrupt truncation.

\textbf{Learned Refinement.} A lightweight single-layer Transformer refiner performs constrained cross-attention from the subsequent segment onto the preceding one. This gently corrects residual mismatches while keeping changes minimal—the denoiser, with its deeper architecture and greater capacity, performs the bulk of reconstruction from noise, so the Stitcher avoids large deviations. Final representations are L2-normalized to match the Compressor's expected norm.

During training, ground-truth offsets enable two auxiliary losses: (i) a position regression loss weighted by existence shares in the error range (penalizing misalignment proportionally to regional importance), and (ii) a cosine-similarity overlap reconstruction loss that promotes view invariance. These losses backpropagate through the denoiser and Splitter, encouraging crisp BOS probabilities and more robust denoising.

\subsection{Integration with Hierarchy and Diffusion}
The Stitcher enforces \emph{hierarchical invariance} by ensuring that overlapping regions produce nearly identical representations after denoising and refinement, thereby stabilizing multi-level abstraction. Stitched level-$l$ outputs are fed directly into level-$l+1$ splitting, supporting unbounded recursive hierarchies.

In DDMM-based generation, all available segments are stitched into a full document representation. For the tail segment, output is truncated where existence probabilities (inferred from BOS predictions) fall below a threshold, preventing padded or low-quality extensions beyond reliable content.

By remaining lightweight yet fully differentiable, and by tightly integrating with Probabilistic Attention and existence probabilities, the Segment Stitcher completes Zonkey's gradient-based pipeline—distinguishing it from segmented methods that trade end-to-end optimization for scalability.

\section{End-to-End Training and Objectives}\label{sec:training}
Our training procedure optimizes Zonkey end-to-end across its hierarchical levels, progressively stabilizing lower representations before advancing to higher abstractions. This curriculum ensures that foundational elements, such as character n-grams, solidify prior to learning sentence-like compressions, promoting emergent linguistic hierarchies without explicit supervision.
The training loop processes batches of raw documents. At each active level $l$, inputs are the stitched outputs from level $l-1$ (or character embeddings at level 0). The forward pass proceeds through the Segment Splitter (producing overlapping segments with probabilistic BOS and existence probabilities), Compressor (yielding fixed-dimensional latents), noise perturbation (per DDMM schedule), Denoiser (reconstructing segments), and Stitcher (reassembling full sequences with overlap refinement). All position-wise losses are reweighted by normalized existence shares from the Splitter, ensuring uniform contribution across overlapping positions regardless of redundancy.
The total loss is a weighted sum across active levels:
$$\mathcal{L} = \sum_{l} w_l \mathcal{L}_l,$$
where $w_l$ prioritize lower levels early in training, and $\mathcal{L}_l$ aggregates the following components (all weighted by existence shares where applicable):

\textbf{Reconstruction Losses.} Contrastive cosine similarity (with in-batch negatives and $\mathrm{atanh}$ transformation) between denoised segments and ground-truth inputs. We compute two variants:
\textbf{Clean}: after minimal final-step noise and denoising—training precise recovery akin to a denoising autoencoder~\citep{vincent2008extracting}.
\textbf{Dirty}: after large accumulated noise (simulating multi-step diffusion via Gaussian perturbations~\citep{song2021denoising}) and denoising—training robust reversal of intermediate perturbations, enabling DDMM's dynamic balance between cautious small steps and ambitious leaps.

\textbf{Collapse Prevention Losses.} Auxiliary cosine similarity penalty on compressed latents (before and after perturbation) to prevent mode collapse. This encourages near-zero cosine similarity between representations from different documents, directly penalizing unwanted correlations and promoting distinct, robust latents~\citep{chen2020simple}.

\textbf{MLM Loss} (all levels except 0). Masked prediction with contrastive reconstruction, clustering semantically similar contexts and guiding the Splitter toward low-entropy linguistic boundaries~\citep{devlin2019bert}.

\textbf{Token Loss} (level 0 only). Cross-entropy on character predictions from decompressed embeddings, grounding the hierarchy in exact lexical recovery.

\textbf{Splitter Regularization Losses}. These auxiliary objectives guide the Segment Splitter toward adaptive, meaningful splits by regularizing BOS probabilities and segment lengths. The primary BOS cross-entropy loss encourages accurate prediction of sequence starts, treating it akin to a reconstruction task where the model "guesses" linguistically plausible boundaries (e.g., at spaces or punctuation) based on contextual embeddings, without explicit labels. Complementing this, a penalty on the average BOS probability discourages overly frequent splits, promoting longer segments for efficient hierarchical compression---for instance, favoring a perfect reconstruction from 6 word-like vectors over 5, as it indicates superior information packing and abstraction. Finally, explicit penalties for excessively short or long segments provide strong safeguards, imposing large losses for violations of computational assumptions (e.g., segments shorter than a minimum threshold or exceeding maximum sequence lengths), ensuring stable and reliable splitting during training and inference.

\textbf{Stitcher Losses.} Weighted offset regression (MSE proportional to existence shares in mismatched regions) and cosine similarity on overlap reconstructions, enforcing hierarchical invariance and seamless reassembly~\citep{song2024hierarchical}.
These objectives collectively drive emergent linguistic structure: reconstruction and MLM ensure fidelity and semantics; diffusion-specific losses enable stable variable-length generation; splitter penalties yield adaptive, meaningful tokenization; and stitcher losses guarantee global consistency. The result is a fully differentiable hierarchy that aligns compressions and splits with downstream reconstruction quality, outperforming heuristic approaches~\citep{yang2024learnable} in domain adaptation and scalability. Empirical training on Wikipedia yields coherent, multi-sentence generations with word- and sentence-level abstractions emerging without explicit supervision.

These objectives collectively drive emergent linguistic structure: reconstruction and MLM ensure fidelity and semantics; splitter penalties yield adaptive, meaningful segmentation; and stitcher losses guarantee global consistency. The result is a fully differentiable hierarchy that aligns compressions and splits with downstream reconstruction quality. Empirical training on Wikipedia yields coherent generation with word-level and sentence-level abstractions emerging without explicit supervision.

\section{Generation and Applications}\label{sec:generation}

Zonkey's hierarchical diffusion framework enables coherent text generation from noise and supports innovative applications such as non-sequential infilling of missing sections in partially complete texts. A key advantage of the hierarchical design is its potential for efficient, scalable generation: unlike autoregressive character-level models or iterative diffusion models that process tokens sequentially, Zonkey generates all vectors at a given hierarchical level in parallel. This allows simultaneous decompression and denoising across entire sentences or paragraphs once higher-level latents are available, offering substantial speed benefits for long-form outputs as the hierarchy deepens.

While the current prototype is trained on a single GPU with Wikipedia data and demonstrates coherent sentence-level generation, scaling to deeper hierarchies (paragraph- or document-level) and larger datasets remains future work limited by computational resources. Quantitative comparisons to existing models (e.g., perplexity or generation metrics) are challenging due to Zonkey's lack of a fixed vocabulary, continuous latent space, and hierarchical output structure, which make standard token-based metrics inapplicable. We therefore focus on qualitative evidence: emergent linguistic hierarchies, coherent variable-length outputs, and adaptive segmentation without explicit supervision.

\subsection{Diffusion-Based Text Generation}

Unconditional text generation in Zonkey proceeds via iterative denoising in the compressed latent space, producing sequences of arbitrary length without fixed tokens or explicit end-of-sequence markers.

At level $l$, the process begins with an initial compressed latent $\mathbf{z}_0 \in \mathbb{R}^{C \times d}$, where $C$ is the number of compression vectors and $d$ is the model dimension. For unconditional generation, $\mathbf{z}_0$ is sampled from $\mathcal{N}(0, I)$ and normalized. Over $T$ diffusion steps with a linear noise schedule decreasing from $\sigma_{\max}$ to $\sigma_{\min}$, the following operations are performed:

\begin{enumerate}
  \item Decompress $\mathbf{z}_t$ into a sequence $\mathbf{x}_t$ using the downward (autoregressive) transformer, conditioned on the current noise level $\sigma_t$.
  \item Denoise $\mathbf{x}_t$ with the bidirectional transformer encoder (using Probabilistic Attention) to obtain $\hat{\mathbf{x}}_t$.
  \item Predict BOS probabilities on $\hat{\mathbf{x}}_t$ to derive existence probabilities $p_{\text{exist}}$, enabling soft truncation.
  \item Recompress $\hat{\mathbf{x}}_t$ (weighted by $p_{\text{exist}}$) to produce $\mathbf{z}_{t-1}$ for the next iteration.
\end{enumerate}

The loop continues until $\sigma_T \approx 0$. The final sequence is truncated where $p_{\text{exist}} < \varepsilon$ (e.g., 0.1). Hierarchical extension is straightforward: denoised outputs from level $l$ can be stitched and fed as input to level $l+1$, enabling progressive generation of multi-sentence or longer text. Preliminary results on Wikipedia data yield coherent, wiki-like sentences with emergent word- and sentence-level structure.

Because all compression vectors at a given level are processed in parallel, generation time grows favourably with sequence length compared to strictly sequential alternatives—making the approach particularly promising for scalable long-form synthesis at deeper hierarchies.

\subsection{Infilling and Targeted Reconstruction}

A major motivation for Zonkey's design is its natural support for infilling arbitrary gaps in existing text, a task at which autoregressive models struggle due to their left-to-right constraint.

For example, at level 1 (sentence-like), given a prefix ``Bob is a'' and suffix ``player in the NBA,'', we compress both into fixed word-like vectors. We then initialise a noisy latent sequence of sufficient length, replacing the prefix and suffix positions with their clean compressed vectors and setting their existence probabilities to 1, while post suffix vectors receive existence probabilities of 0. During denoising:

\begin{enumerate}
  \item Decompress the full latent, keeping known prefix/suffix vectors fixed.
  \item Denoise selectively—updating only the noisy middle region while preserving fixed regions.
  \item Recompress with existence weighting, propagating refinements.
\end{enumerate}

Over iterations, the gap converges to a coherent infill such as ``basketball,'' producing ``Bob is a basketball player in the NBA.''

This mechanism scales naturally to higher levels (e.g., infilling missing paragraphs or chapters given surrounding context) once deeper hierarchies are trained. The combination of fixed-vector conditioning, probabilistic existence handling, and differentiable stitching makes non-sequential completion particularly effective.

\subsection{Limitations and Future Work}

The current implementation is a proof-of-concept trained on limited resources, reaching reliable sentence-level coherence (levels 0--1). Deeper hierarchies and document-scale generation require substantially more compute and data, which we leave to future work. Similarly, systematic quantitative evaluation (e.g., human ratings, reconstruction fidelity metrics, or domain-adaptation benchmarks) is planned for larger-scale versions.

In summary, Zonkey demonstrates that fully differentiable hierarchical diffusion models can produce coherent text with emergent linguistic structure. Its efficiency advantages and native support for infilling position it as a promising direction toward scalable, adaptable, and truly gradient-based language models.

\bibliography{custom}
\end{document}